\documentclass{article}
\usepackage{times}
\usepackage{algorithm,algorithmic,a4wide,amssymb}
\begin{document}
\def\GM{G\"{o}del Machine }
\def\gm{G\"{o}del machine }
\def\GMn{G\"{o}del Machine}
\def\gmn{G\"{o}del machine}
\def\hs{Hsearch }
\date{}

\title{Self-Delimiting Neural Networks \\ 
{\small Technical Report IDSIA-08-12} }

\date{June 2012, revised September 2012}
\author{J\"{u}rgen Schmidhuber  \\
The Swiss AI Lab IDSIA \\
Galleria 2, 6928 Manno-Lugano, Switzerland \\
University of Lugano \& SUPSI, Switzerland}
\maketitle

\begin{abstract}
Self-delimiting (SLIM) programs are a central concept of theoretical computer science, particularly 
algorithmic information \& probability theory, and asymptotically optimal program search (AOPS). 
To apply AOPS to  (possibly recurrent) neural networks (NNs),
I introduce SLIM NNs.
A typical SLIM NN is a general parallel-sequential computer.
Its neurons have threshold activation functions.
Its output neurons may affect the  environment, which may respond with new inputs.
During a computational episode,
activations are spreading from input neurons through the SLIM NN
until the computation activates a special {\em halt neuron}.
Weights of the NN's used connections define its program.
Halting programs form a prefix code.
An episode may never activate most neurons,
and hence never even consider their outgoing connections.
So we trace only neurons and connections used at least once.
With such a trace,
the reset of the initial NN state does not cost more than the latest program execution.
This by itself may speed up traditional NN implementations.
To efficiently change SLIM NN weights based on experience, 
any learning algorithm (LA) should ignore all unused weights.
Since prefixes of SLIM programs influence their suffixes
(weight changes occurring early in an episode influence which weights are considered later),
SLIM NN LAs should execute weight changes online during activation spreading.
This can be achieved by applying AOPS to growing SLIM NNs.
Since SLIM NNs select their own task-dependent effective size ($=$number of used free parameters),
they have a built-in way of addressing overfitting, 
with the potential of effectively becoming small and {\em slim} whenever this is beneficial.
To efficiently teach a SLIM NN  to solve many tasks,
such as correctly classifying many different patterns,
or solving many different robot control tasks,
each connection keeps a list of tasks it is used for.
The lists may be efficiently updated during training.
To evaluate the overall effect of currently tested weight changes,
a SLIM NN LA needs to re-test performance only on 
 the efficiently computable union of tasks potentially affected by the current weight changes.
Search spaces of many existing LAs (such as hill climbing and neuro-evolution) can be greatly reduced
by obeying  restrictions of SLIM NNs. 
Future SLIM NNs will be implemented on 3-dimensional brain-like multi-processor hardware.
Their LAs will minimize task-specific total wire length of used connections,
to encourage efficient solutions of subtasks by subsets of neurons that are physically close.
The novel class of SLIM NN LAs is currently being probed in ongoing experiments
to be reported in separate papers.

\end{abstract}

\newpage

\section{Traditional NNs / Motivation of SLIM NNs / Outline}
\label{motivation}

Recurrent neural networks (RNNs) are neural networks (NNs)  \cite{bishop:2006}
with feedback connections. RNNs are, in principle,
as powerful as any traditional computer.
There is a trivial way of seeing
this~\cite{Schmidhuber:90thesis}:
a traditional microprocessor can be modeled as
a very sparsely connected RNN consisting of 
simple neurons implementing nonlinear
AND and NAND gates.
Compare~\cite{siegelmann91turing}  for a more complex
argument. 
RNNs can learn to solve many tasks involving
sequences of
continually varying inputs.
Examples include robot control, speech recognition,
music composition, attentive vision, and numerous
others.
Section \ref{intro} will give a brief overview of recent NNs and RNNs
that achieved extraordinary success in many applications and competitions.

Although RNNs are general computers whose programs are weight matrices,
{\em asymptotically optimal program search}  (AOPS)
\cite{Levin:73,Schmidhuber:97bias,Schmidhuber:03nips,Schmidhuber:04oops}
has not yet been applied to RNNs. Instead most RNN learning algorithms are based on
more or less heuristic search techniques such as gradient descent or evolution
(see Section \ref{intro}). 
One reason for the current lack of AOPS-based RNNs 
may be that traditional AOPS variants are designed to search
a space of  {\em sequential} self-delimiting (SLIM) programs
\cite{Levin:74,Chaitin:75}  (Section \ref{sequential}). 
The concept of partially {\em parallel} SLIM NNs will help to adapt AOPS to RNNs.

Section \ref{problems} will mention additional problems addressed by SLIM NNs:
(1) Traditional NN implementations based on matrix 
multiplications may be inefficient for large NN
where most weights are rarely used (Section \ref{inefficiencies}).
(2) Traditional NNs use 
ad hoc ways of avoiding overfitting (Section \ref{overfitting}).
(3)  Traditional RNNs are not well-suited
as increasingly general problem solvers to be trained from scratch by {\sc PowerPlay}
 \cite{Schmidhuber:11powerplay}, which 
continually invents the easiest-to-add novel computational problem by itself
(Section \ref{powerplay}).

Section \ref{rnn} will describe essential properties of SLIM NNs;
Section \ref{universal} will
show how to  apply incremental AOPS to SLIM RNNs.

\subsection{Brief Intro to Successful RNNs and Related Deep NNs  \cite{Schmidhuber:07newmillenniumai}}
\label{intro}

Supervised RNNs can be trained to map sequences of input patterns to desired output sequences
by gradient descent and other methods~\cite{Werbos:88gasmarket,WilliamsZipser:92,RobinsonFallside:87tr,Schmidhuber:92ncn3,Maass:02,Jaeger:04,ICML2011Sutskever_524}. 
Early RNNs had problems with learning to store relevant events in
short-term memory across long time lags
 ~\cite{Hochreiter:01book}. Long Short-Term Memory (LSTM) overcame these problems, outperforming early RNNs in many applications~\cite{Hochreiter:97lstm,Gers:01ieeetnn,Gers:02jmlr,Schmidhuber:06nc,Graves:06icml,graves:08nips,graves:2009nips,Graves:09tpami}.
While RNNs used to be toy problem methods in the 1990s, they have recently started to beat all other methods in 
 challenging real world applications \cite{schmidhuber2011agi,Schmidhuber:06nc,Santi:07ijcai,graves:08nips,Graves:09tpami,graves:2009nips}.
Recently, CTC-trained \cite{Graves:06icml} mulitdimensional \cite{graves:2009nips} RNNs won three Connected Handwriting Recognition Competitions at ICDAR 2009  (see below).

Training an RNN by standard methods is as difficult as training a deep feedforward NN (FNN) with many layers
 \cite{Hochreiter:01book}.
However, recent deep FNNs with special internal architecture overcome these problems to the extent that they are currently winning many international visual pattern recognition contests
\cite{schmidhuber2011agi,ciresan:2010,ciresan:2011ijcai,ciresan:2011ijcnn,ciresan2012cvpr,ciresan:2012NN}
(see  below).
None of this requires the traditional sophisticated computer vision techniques developed over the past six decades or so. Instead, those biologically rather plausible NN architectures learn from experience with millions of training examples.
Typically they have
many non-linear processing stages like  Fukushima's Neocognitron \cite{fukushima:1980}; they sometimes (but not always) profit from sparse network connectivity and techniques such as weight sharing \& convolution \cite{lecun:1998,behnke:2003}, max-pooling \cite{scherer:2010},
and contrast enhancement  like the one automatically
generated by unsupervised {\em Predictability Minimization} \cite{Schmidhuber:92ncfactorial,Schmidhuber:96ncedges,pm}.  NNs are now often outperforming all other methods including the theoretically less general and less powerful 
support vector machines (SVMs) based on statistical learning theory \cite{Vapnik:95} (which for a long time had the upper hand, at least in practice).
These results are currently contributing to a second {\em Neural Network
ReNNaissance} (the first one happened in the 1980s and early 90s) which
might not be possible without  dramatic advances in computational power per Swiss Franc, obtained in the new millennium. In particular,  to implement and train NNs, we exploit graphics processing units (GPUs). GPUs are mini-supercomputers normally used for video games, often 100 times faster than traditional CPUs, 
and a million times faster than PCs of two decades ago when we started this type of research.

Since 2009, my group's NN and RNN methods have achieved many first ranks in international competitions:
{\bf (7)}
ISBI 2012 Electron Microscopy Stack Segmentation Challenge (with superhuman pixel error rate) \cite{ciresan2012nips}. 
{\bf (6)}
IJCNN 2011 on-site Traffic Sign Recognition Competition (0.56\% error rate, the only method better than humans, who achieved 1.16\% on average; 3rd place for 1.69\%) \cite{ciresan:2012NN} 
{\bf (5)}
ICDAR 2011 offline Chinese handwritten character recognition competition \cite{ciresan2012cvpr}. 
{\bf (4)}
Online German Traffic Sign Recognition Contest (1st \& 2nd rank; 1.02\% error rate) \cite{ciresan:2011ijcnn}. 
{\bf (3)}
ICDAR 2009 Arabic Connected Handwriting Competition  (won by  LSTM RNNs \cite{Graves:09tpami,graves:2009nips}, same below). 
{\bf (2)}
ICDAR 2009 Handwritten Farsi/Arabic Character Recognition Competition. 
{\bf (1)}
ICDAR 2009 French Connected Handwriting Competition.
Additional
 1st ranks were achieved in important machine learning (ML) benchmarks since 2010: 
{\bf (A)} MNIST handwritten digits data set   \cite{lecun:1998} (perhaps the most famous ML benchmark). New records: 0.35\% error in 2010 \cite{ciresan:2010}, 0.27\% in 2011 \cite{ciresan:2011icdar}, first human-competitive performance (0.23\%) in 2012 \cite{ciresan2012cvpr}. 
{\bf (B)} NORB stereo image data set \cite{lecun:2004}. New records in 2011, 2012, e.g., \cite{ciresan2012cvpr}. 
{\bf (C)} CIFAR-10 image data set  \cite{krizhevsky:2009}. New records 
in 2011, 2012, e.g., \cite{ciresan2012cvpr}.

{\em Reinforcement Learning} (RL) \cite{Kaelbling:96,Sutton:98} is more challenging than
supervised learning as above, since there is no teacher
providing desired outputs at appropriate time steps.
To solve a given problem, the learning agent itself
must discover useful output sequences in response
to the observations.
The traditional approach to RL  \cite{Sutton:98} makes strong assumptions
about the environment, such as the Markov assumption:
the current input of the agent tells it all it needs to know
about the environment.
Then all we need to learn is some sort of reactive mapping from
stationary inputs to outputs. This is often unrealistic.
A more general approach for partially observable environments
directly evolves programs for RNNs
with internal states  (no need
for the Markovian assumption), by applying evolutionary
algorithms \cite{Rechenberg:71,Schwefel:74,Holland:75} to RNN weight matrices
\cite{yao:review93,Sims:1994:EVC,stanley:ec02,hansen2001ecj}.
Recent work brought progress through a focus on reducing
search spaces by co-evolving the comparatively small
weight vectors of individual 
 neurons and synapses~\cite{Gomez:08jmlr}, by Natural Gradient-based Stochastic Search Strategies~\cite{wierstraCEC08,Sun2009a,Sun2009,Schaul2010pybrain,glasmachers:2010b,wierstra2010}, and by reducing search spaces through weight matrix compression \cite{Schmidhuber:97nn,koutnik:gecco10}.
 RL RNNs now outperform many previous methods on benchmarks~\cite{Gomez:08jmlr}, creating memories of important events and solving numerous tasks unsolvable by classical RL methods.

\subsection{Principles of Traditional Sequential SLIM Programs}
\label{sequential}

The RNNs of Section \ref{intro} are not designed for AOPS.
Traditional AOPS favors {\em short and fast} programs written in a universal 
programming language that permits
 {\em self-delimiting} (SLIM) programs
 \cite{Levin:74,Chaitin:75} 
studied in the theory of Kolmogorov complexity and algorithmic 
probability  \cite{Solomonoff:64,Kolmogorov:65,Schmidhuber:95kol,LiVitanyi:97,Schmidhuber:97nn,Schmidhuber:02ijfcs,Schmidhuber:02colt,Hutter:05book+}. In fact, SLIM programs are essential for making the
theory of algorithmic probability elegant, e.g.,  \cite{LiVitanyi:97}.

The nice thing about SLIM programs is that they determine their own size during runtime.
Traditional {\em sequential} SLIM programs work as follows: Whenever  the instruction pointer of a Turing Machine or a traditional PC 
 has been initialized or changed (e.g., through a conditional jump instruction) such that its new value points to an address containing some executable instruction, then  the instruction will be executed. This may change the internal storage including the instruction pointer. 
 Once a halt instruction is encountered and executed, the program stops. 
  
 Whenever the instruction pointer points to an address that never was used before by the current program
 and does not yet contain an instruction, this is interpreted as the online request for a new instruction \cite{Levin:74,Chaitin:75}
(typically selected by a time-optimal search algorithm \cite{Schmidhuber:97bias,Schmidhuber:03nips,Schmidhuber:04oops}).
The new instruction is appended to the growing list of used instructions defining the program so far.

Executed program beginnings or {\em prefixes} influence their possible suffixes.
Code execution determines code size in an online fashion. 

Prefixes that halt or at least cease to request any further input instructions are called self-delimiting programs or simply programs. This procedure yields {\em prefix codes} on program space. No halting or non-halting program can be the prefix of another one. 

Principles of SLIM programs are not implemented by traditional standard RNNs.
SLIM RNNs, however,  do implement them,
making SLIM RNNs highly compatible with time-optimal program search (Section \ref{universal}).

\subsection{Additional Problems of Traditional NNs Addressed By Slim NNs}
\label{problems}

\subsubsection{Certain Inefficiencies of Traditional NN Implementations}
\label{inefficiencies}
Typical matrix multiplication-based implementations of the  NN algorithms in Section \ref{intro}  {\em always} take into 
consideration {\em all} neurons and connections of a given NN, 
even when most are not even needed to solve a particular task. 

The SLIM NNs of the present paper use more efficient ways of information 
processing and learning.
Imagine a large RNN with a 
trillion connections connecting a
billion neurons, each with a thousand outgoing connections to other neurons.
If the RNN consists of biologically plausible winner-take-all (WITA) 
neurons with threshold activation functions \cite{Schmidhuber:89cs},
also found in networks of spiking neurons \cite{gerstnerbook},
a given RNN computation might activate just a tiny fraction of all neurons,
and hence never even consider the outgoing connections of most neurons.
This simple fact can be exploited to devise classes of 
NN algorithms that are less costly in various senses, to be detailed below.

\subsubsection{Traditional Ad Hoc Ways of Avoiding Overfitting}
\label{overfitting}
To avoid overfitting on training sets
and to improve generalization on test sets, 
various pre-wired regularizer terms  \cite{bishop:2006} have been added to 
performance measures or objective functions of traditional NNs.
The idea is to obtain {\em simple} NNs by penalizing NN complexity. 
One problem is the {\em ad hoc} weighting of such additional terms.
The present paper's more principled  SLIM NNs can  learn to actively select
in task-specific ways their 
own size, that is, their effective number of weights ($=$ modifiable free parameters),
in line with the theory of  algorithmic 
probability and optimal universal inductive inference
 \cite{Solomonoff:64,Kolmogorov:65,Schmidhuber:95kol,LiVitanyi:97,Schmidhuber:97nn,Schmidhuber:02ijfcs,Schmidhuber:02colt,Hutter:05book+}.

\subsubsection{RNNs as Problem Solvers for PowerPlay}
\label{powerplay}

The recent unsupervised {\sc PowerPlay} framework \cite{Schmidhuber:11powerplay} trains an 
increasingly general problem solver from scratch, 
continually inventing the easiest-to-add novel computational problem by itself.
We will see that unlike traditional RNNs, SLIM RNNs are well-suited
as problem solvers to be trained by {\sc PowerPlay}. In particular, SLIM RNNs support 
a natural modularization of the space of self-invented and other tasks and their solutions 
into more or less independent regions. More on this particular motivation
of SLIM NNs can be found in Section \ref{experiments}.

\section{Self-Delimiting Parallel-Sequential Programs on SLIM RNNs}
\label{rnn}


Unless stated, or otherwise obvious,
to simplify notation,
throughout the paper newly introduced variables
are assumed to be integer-valued and to cover the range implicit in the context.
 $\mathbb{N}$ denotes the natural numbers,
 $\mathbb{R}$ the real numbers,
 $\epsilon \in \mathbb{R}$ a positive constant,
  $m,n,n_0,k,i,j,l,p,q$ non-negative integers.
 
The $k$-th computational unit or {\em neuron} of our RNN is denoted $u^k$
($0<k \leq n(u) \in \mathbb{N}$).
 $w^{lk}$ is the real-valued {\em weight} on the directed connection  $c^{lk}$
from $u^l$ to $u^k$. 
Like the human brain, the RNN may be sparsely connected, that is, 
each neuron may be connected to just a fraction of the other neurons.
To {\em program} the RNN means to set some or all of the weights $\langle w^{lk} \rangle$.


At discrete time step $t=1,2,\ldots,t_{end}$ of a finite interaction sequence with the environment (an episode),
$u^k(t)$ denotes the real-valued {\em activation} of $u^k$.
The  real-valued input vector $x(t)$ (which may include a unique encoding of the current task)
has $n(x) \in \mathbb{N}$ components, where
the $k$-th component is denoted $x^k(t)$;
we define $u^k(t)=x^k(t)$ for $k=1,2,\ldots, n(x)$.
That is, the first $n(x)$ neurons are input neurons;
they do not have incoming connections from other neurons.
The current reward signal $r(t)$ (if any) is a special real-valued input; 
we set $u^{n(x)}(t)=r(t)$. 
For $k=n(x)+1,\ldots, n(x)+n(y)$, we set $y^k(t)=u^k(t)$,
thus defining the $n(y)$-dimensional output vector $y(t)$, which 
may affect the environment (e.g., by defining a robot action) and thus  future $x$ and $r$.
For $n(x)+1<k \leq n(u)$ we initialize $u^k(1)=0$ and for $1 \leq t < t_{end}$
compute $u^k(t+1)=f^k(\sum_l w^{lk}u^l(t))$ (if $u^k$ is user-defined as an {\em additive} neuron)
or $u^k(t+1)=f^k(\prod_l w^{lk}u^l(t))$  (if $u^k$ is a {\em multiplicative} neuron).

Here the function $f^k$ maps $\mathbb{R}$ to $\mathbb{R}$. 
Many previous RNNs  use differentiable activation functions such as
$f^k(x)=1/(1+e^{-x})$, or $f^k(x)=x$.
We want SLIM NN programs that can easily define their own size.
Hence we focus on {\em threshold activation functions} that allow for keeping most units inactive
most of the time, e.g.: $f^k(x)=1$ if $x \geq 0.5$ and 0 otherwise.
For the same reason
we also consider {\em winner-take-all activation functions}.
Here all non-input neurons (including output neurons)
are partitioned into ordered winner-take-all subsets (WITAS),
like in my first RNN from 1989 \cite{Schmidhuber:89cs}.
Once all $u^k(t+1)$ of a WITAS are computed as above, 
and at least one of them exceeds a threshold such as 0.5,
and a particular $u^k$ is the first with maximal activation in its WITAS,
then we re-define $u^k(t+1)$ as 1, otherwise as 0.

For each  $c^{lk}$  there is a constant cost  $cost^{lk}$ of using  $c^{lk}$ 
between $t$ and $t+1$, provided $w^{lk}u^l(t) \neq 0$. More on this 
in Section \ref{3d}.

A special, unusual, non-traditional 
 non-input neuron is called the {\em halt neuron} $u^{halt}$.
If $u^{halt}$ is active (has non-zero activation)  once all updates of time $t$ have been completed, we define $t_{end}:=t$, and the computation stops.
For non-halting programs, $t_{end}$ might be a maximal time limit $t_{lim}$ to be defined
by a learning algorithm 
based on techniques of asymptotically 
optimal program search \cite{Levin:73,Schmidhuber:97bias,Schmidhuber:03nips,Schmidhuber:04oops}---see Section \ref{universal}.

\subsection{Efficient Activation Spreading and NN Resets}
\label{spreading}

\begin{algorithm} {\bf Procedure \ref{spread}: Spread}
\label{spread}
\begin{algorithmic}
\STATE (see text for global variables and their initialization before the first call of {\bf Spread})
\STATE set $new:=old:=trace:=nil$
\WHILE {$u^{halt}(now) <$  threshold}
\STATE get next input vector $x(now)$ 
\FOR {$k=1, 2,\ldots, n(x)$} 
\STATE $u^k(now):=x^k(now)$; if $u^k(now)\neq 0$ append $u^k$ to $old$
\ENDFOR
\FOR {all $u^l \in old$} 
     \FOR {all $c^{lk} \in out^l$} 
       \STATE {\bf (*)} [If  $c^{lk}$ was never used before in the current or any previous episode, a learning algorithm (see Section \ref{learning}) may set  $w^{lk} \neq 0$ for the first time, thus growing the effectively used SLIM RNN by  $c^{lk}$ (and by $u^k$ in case $u^k$ was never used before)]
       \IF {$w^{lk} \neq 0$}
       \STATE if $mark^{lk} =0$ then set $mark^{lk}:=1$ and append  $c^{lk}$ to $trace$
       \STATE if $u^k$ is {\em additive} then $next^k:=next^k+u^l(now)w^{lk}$
       \STATE else $u^k$ is {\em multiplicative} and $next^k:=next^k u^l(now)w^{lk}$
       \STATE if $used^k=0$ then set $used^k:=1$ and append $u^k$ to $new$
       \STATE $time:=time + cost^{lk}$ [long wires may cost more---see Section \ref{3d}]
       \STATE (**) if $time > t_{lim}$ then exit {\bf while} loop
       \ENDIF
       \ENDFOR
\ENDFOR
\FOR {$u^l \in new$} 
\STATE determine final new activation $u^l(now)$ (either 1 or 0)
through thresholding and determination of WITAS winners (if any; see Section \ref{rnn})
\STATE   $used^l:=0$; if $u^l$ is  {\em additive} then $next^l:=0$ else $next^l:=1$ [restore]
\ENDFOR
\STATE $old := new$; $new:=nil$ [now $old$ cannot contain any input units]
\STATE delete from $old$ all $u^l$ with zero $u^l(now)$
\STATE execute environment-changing actions (if any) based on output neurons;
possibly update problem-specific variables needed for  an ongoing performance evaluation according to a given
problem-specific objective function \cite{bishop:2006} (see Section \ref{learning} on learning);
continually add the computational costs of the above to $time$; once $time > t_{lim}$ exit {\bf while} loop
\ENDWHILE
\FOR {$u^l \in new$ [perhaps $new \neq nil$ in case of premature exit from (**)]} 
\STATE   $used^l:=0$; if $u^l$ is  {\em additive} then $next^l:=0$ else $next^l:=1$ [restore]
\ENDFOR
\FOR {$c^{lk} \in trace$} 
\STATE  $mark^{lk}:=0$
\ENDFOR
\end{algorithmic}
\end{algorithm}

Procedure {\bf Spread} (inspired by an earlier RNN implementation \cite{Schmidhuber:89cs})
efficiently implements episodes according to the formulae above (see Procedure \ref{spread}).
Each $u^k$ is associated with a list $out^k$ of all connections emanating from $u^k$.
A nearly trivial observation is that 
only neurons with non-zero activation need to be considered for activation spreading.
There are three global variable lists (initially empty): $old$, $new$, $trace$.
Lists $old$ and $new$ track neurons used in the most recent two time steps, 
to efficiently proceed from one time step to the next;
$trace$ tracks connections used at least once during the current episode.
For each $u^l$ there is a global Boolean variable $used^l$ (initially 0),
to mark which RNN neurons already received contributions from the previous step 
during the current interaction sequence with the environment.
For each $c^{lk}$ there is a global Boolean variable $mark^{lk}$  (initially 0),
to mark which connections were used at least once.
The following real-valued variables are initalized by 0 unless indicated otherwise:
$u^k(now)$ holds the activation of $u^k$ at the current step,
$next^k$ is a temporary variable for collecting contributions from neurons connected to $u^k$ 
(initialized by 1 if $u^k$ is a {\em multiplicative} neuron);
$x^k(now)$ holds the current input of $u^k$ if $u^k$ is an input neuron.
The integer variable $time$ (initially 0) is used to count connection usages;
the given time limit $t_{lim}$ will eventually stop episodes  that are not halted
by the halting unit. The label {\bf (*)}  in  {\bf Spread} will be referred
to in Section \ref{learning} on learning.
{\bf Spread}'s results include two global variables: the program $trace$ and its runtime $time$.

Once {\bf Spread} has finished, weights of connections in $trace$
are the only {\em used instructions} of the SLIM program that just ran on the RNN.
We observe: tracking and undoing the effects of a program
essentially does not cost more than its execution, because untouched parts of the net 
are never considered for resets.

Note the difference to most standard NN implementations: the latter use matrix multiplications
to multiply entire weight matrices by activation vectors. The simple list-based method {\bf Spread},
however, ignores all unused neurons and irrelevant connections. In large brain-like sparse nets this by itself
may dramatically accelerate information processing.

\subsection{Relation to Traditional Self-Delimiting Programs and Prefix Codes}
\label{programs}

Since the order of neuron activation updates between two successive time steps is irrelevant,
such updates can be performed in parallel. That is, 
SLIM NN code can be executed in partly parallel and partly sequential fashion.
Nevertheless, the execution follows basic principles of
sequential SLIM programs \cite{Levin:74,Chaitin:75,LiVitanyi:97,Schmidhuber:02ijfcs,Schmidhuber:02colt,Hutter:05book+}
(Section \ref{sequential}).

As mentioned in Section \ref{sequential}, the latter form a  {\em prefix code} on program space.
An equivalent condition holds  for the $trace$s computed by {\bf Spread} in a resettable deterministic environment,
as long as we identify a given $trace$ with all possible $trace$ variants (an equivalence class) reflecting
the irrelevant order of neuron activation updates between two successive time steps.
(In non-resettable environments, the environmental inputs have to be viewed as 
an additional part of the program
to establish such a prefix code condition.)
Compare also Section \ref{growth} on learning-based NN growth and the label {\bf (*)}  in  {\bf Spread}.

\section{Principles of Efficient Learning Algorithms (LAs) for SLIM NNs}
\label{learning}

Through weight changes,
the NN is supposed to learn something from a sequence of training task descriptions $T_1, T_2, \ldots$. 
Here each unique $T_i  \in \mathbb{R}^{n(T_i)}$ $(i=1, 2, \ldots)$ 
could identify a pattern classification task or robot control task,
where the task description dimensionality $n(T_i)$ is an integer constant,
such that (parts of) $T_i$ can be used as a non-changing part of the inputs $x(t)$.
The SLIM NN's performance on each task is measured by some given
problem-specific objective function \cite{bishop:2006}. 

To efficiently change SLIM NN weights based on experience, 
any learning algorithm (LA) should ignore all unused weights.

Since prefixes of SLIM programs influence their suffixes,
and weights used early in a {\bf Spread} episode influence which weights are considered later,
weight modifications tested by SLIM NN LAs should be generated online during  program
execution, such that unused weights are not even considered as candidates for change.
Search spaces of many well-known  LAs (such as hill climbing and neuro-evolution; see Section \ref{intro}) 
obviously may be greatly reduced
by obeying these restrictions.

\subsection{ LA-Based SLIM NN Growth}
\label{growth}

Typical SLIM NN LAs (e.g., Section \ref{universal})
will influence how SLIM NNs grow.
Consider the bracketed statement in procedure
 {\bf Spread} labeled by {\bf (*)}.
If some $c^{lk}$ considered here was never used before,
and its $w^{lk}$ never defined, 
a tentative value  $w^{lk} \neq 0$ can be temporarily set here (setting $w^{lk} = 0$ wouldn't have any effect),
and the used part of the net effectively grows by  $c^{lk}$ (and $u^k$ in case $u^k$ also was never used before).
Later performance evaluations may suggest to make this extended topology permanent 
 and keep  $w^{lk}$ as a basis for further changes.

This type of SLIM program-directed NN growth is quite different from previous 
popular NN growth strategies, e.g., \cite{fritzke94}.

\subsection{(Incremental Adaptive) Universal Search for SLIM NNs}
\label{universal}

LAs for growing SLIM NNs as in Section \ref{growth} may be based on techniques of asymptotically 
optimal program search \cite{Levin:73,Schmidhuber:97bias,Schmidhuber:03nips,Schmidhuber:04oops}.
Assume some initial bias in
form of  probability distributions $P^{lk}$ on a finite set $V ={v_1,v_2,\ldots,v_m}$ of possible 
real-valued values 
for each $w^{lk}$. 
Let  $n(c^{lk})$ denote the number of usages of $c^{lk}$ during {\bf Spread}.
Given some task, one of the simplest LAs 
based on {\em universal search} \cite{Levin:73} 
is the algorithm {\bf Universal SLIM NN Search}.

\begin{algorithm}{\bf Universal SLIM NN Search} (Variant 1)
\label{fast}
\begin{algorithmic}
\FOR {$i := 1, 2, \ldots$} 
\STATE  systematically enumerate and test possible programs $trace$ (as computed by {\bf Spread}) with runtime \\
$ExternalCosts(trace) + \sum_{c^{lk} \in trace} cost^{lk} n(c^{lk}) \leq 2^i \prod_{c^{lk} \in trace} P^{lk}(w^{lk})$ \\
until all have been tested, or the most recently tested $trace$ has solved the task and the solution has been verified; in the latter case exit and return that $trace$
\ENDFOR
\end{algorithmic}
\end{algorithm}

Here the real-valued expression $ExternalCosts(trace)$ represents all 
costs other than those of the NN's connection usages. This
includes the costs of output actions and evaluations.
$ExternalCosts(trace)$ may be negligible in many applications though.
The left-hand side of the inequality in {\bf Universal SLIM NN Search}
 is essentially the $time$ computed by  {\bf Spread}.

That is, {\bf Universal SLIM NN Search} time-shares all program tests such that each program $trace$ gets not more than a
constant fraction  of the total search time. This fraction is proportional to its probability.
The method is near-{\em bias-optimal} \cite{Schmidhuber:04oops} and
 {\em asymptotically optimal} in the following sense:
If some unknown $trace$
requires at most $f(k)$ steps to solve a problem of a given class and integer size $k$ and verify the
solution, 
where $f$ is a computable function mapping integers to integers, then
the entire search also will need at most $O(f(k))$ steps.

To explore the space of possible $trace$s and their computational effects,
efficient implementations of {\bf Universal SLIM NN Search} use  depth-first search in program prefix space 
combined with stack-based backtracking for partial state resets,
like in the online source code \cite{Schmidhuber:04oopscode}
of 
 the {\em Optimal  Ordered Problem Solver OOPS}
 \cite{Schmidhuber:04oops}.

Traditional NN LAs address overfitting by pre-wired regularizers and hyper-parameters \cite{bishop:2006}
to penalize NN complexity. {\bf Universal SLIM NN Search}, however, systematically tests
programs that essentially select their
 own task-dependent size (the number of weights $=$ modifiable free parameters).
It favors SLIM NNs that combine short runtime and simplicity/low descriptive complexity. Note  that 
small size or low description length is equivalent to high probability, since the negative binary logarithm of the probability of some SLIM NN's $trace$ is essentially the number of bits needed to encode  $trace$ by Huffman coding \cite{Huffman:52}.
Hence the method has  a built-in way of addressing overfitting and boosting generalization performance
\cite{Schmidhuber:95kol,Schmidhuber:97nn} through a bias towards simple solutions
in the sense of Occam's razor \cite{Solomonoff:64,Kolmogorov:65,LiVitanyi:97,Hutter:05book+}.

The method can be extended \cite{Schmidhuber:97bias,Schmidhuber:04oops} such that it {\em incrementally} 
solves each problem in an ordered  {\em sequence} of problems,
continually organizing and managing and reusing earlier acquired knowledge.
For example,
this can be done by updating
the probability distributions $P^{lk}$  based on success: once {\bf Universal SLIM NN Search}{\bf Universal SLIM NN Search} has found a solution $trace$
to the present problem,  some (possibly heuristic) strategy is used to {\em shift the bias} by increasing/decreasing
the probabilities of weights of connections in $trace$ 
before the next invocation of {\bf Universal SLIM NN Search}
on the next problem \cite{Schmidhuber:97bias}.
Roughly speaking, each doubling of $trace$'s probability halfs the time needed by {\bf Universal SLIM NN Search} to find $trace$.

One of the simplest bias-shifting procedures is 
{\bf Adaptive Universal SLIM NN Search}  (Variant 1) based on earlier 
work on sequential programs \cite{Schmidhuber:97bias}.
It uses a constant learning rate $\eta \in \mathbb{R}; 0 < \eta <1$. 
After a successful episode with a halting program, for each $c^{lk}$ with $n(c^{lk})>0$, 
let $yes^{lk}$ denote how often successive activations $u^l(t)$ and $u^k(t+1)$ $(t<t_{end})$ were both 1,
and let $no^{lk}$ denote how often $u^l(t)$ was 1 but $u^k(t+1)$ was 0.
Note that $n(c^{lk})=yes^{lk}+no^{lk}$. Define $-1 \leq \triangle^{lk}:=(yes^{lk}-no^{lk})/n(c^{lk}) \leq 1$.
The sign of $\triangle^{lk}$ indicates whether $u^l$ usually helped to trigger or suppress $u^k$.
A Hebb-inspired learning rule uses $\triangle^{lk}$ to change $P^{lk}$ in case of success.

\begin{algorithm}{\bf Adaptive Universal SLIM NN Search}  (Variant 1)
\label{adaptive}
\begin{algorithmic}
\FOR {$i := 1, 2, \ldots$} 
\STATE use {\bf Universal SLIM NN Search} to  solve the $i$-th problem by some solution program  $trace$
\FOR {all $l$ satisfying $c^{lk} \in trace$}
\FOR {all $c^{lk} \in out^l$}
\IF {$\triangle^{lk}<0$} 
     \STATE $P^{lk}(w^{lk}):= P^{lk}(w^{lk}) + \eta \triangle^{lk} P^{lk}(w^{lk})$ [decrease $P^{lk}(w^{lk})$]
\ELSE  \STATE $P^{lk}(w^{lk}):= P^{lk}(w^{lk}) + \eta \triangle^{lk}  (1-P^{lk}(w^{lk}))  $  [decrease $1-P^{lk}(w^{lk})$]
\ENDIF
\FOR {all $v \in V, v \neq w^{lk}$} 
\STATE normalize:  $P^{lk}(v):=\gamma P^{lk}(v)$, where constant $\gamma \in \mathbb{R}$ is chosen to ensure $\sum_j P^{lk}(v_j)=1$
\ENDFOR
\ENDFOR
\ENDFOR
\ENDFOR
\end{algorithmic}
\end{algorithm}

To reduce the search space,
alternative (adaptive) {\bf Universal SLIM NN Search} variants do not use
independent $P^{lk}$ but joint distributions $P^l$ for each $out^l$
to correlate various $P^{lk}$. For example, the rule may be: exactly one of the connections $\in out^l$ must have a weight  of 1, all others must have -1. (This is inspired by biological brains whose
connections are mostly inhibitory.) The initial $P^l$ may assign equal {\em a priori}
 probability to the  possible weight vectors (as many as there are connections in $out^l$).

Yet additional variants of 
adaptive universal search for low-complexity networks search in 
compressed network space \cite{koutnik:agi10,koutnik:gecco10,ppsn2012cncs}.
Alternatively, apply the principles of the {\em Optimal  Ordered Problem Solver OOPS}
 \cite{Schmidhuber:04oops,Schmidhuber:03nips} to SLIM NNs:
If a new problem can be solved faster by writing a program 
that invokes previously found code
than by solving the new problem from scratch, 
then OOPS will find this out.

\subsection{Tracking which Connections Affect which Tasks}
\label{tracking}

To efficiently exploit that possibly many weights are not used by many tasks,
we keep track of which connections affect which tasks \cite{Schmidhuber:11powerplay}.
For each connection $c^{lk}$  a variable  list $L^{lk}=(T^{lk}_1, T^{lk}_2, \ldots)$ of tasks is introduced. 
Its initial value before learning is $L^{lk}_0$, an empty list.

Let us 
now assume tentative changes of certain used $w^{lk}$ are  computed 
by an LA embedded within a {\bf Spread}-like framework (Section \ref{spreading})---compare 
label {\bf (*)}  in  {\bf Spread}.
That is, 
some of the {\em used} weights (but no unused weights!)
are modified or generated through an LA, 
while it is becoming clear during the ongoing activation spreading computation
which units and connections are used at all---compare Sections \ref{growth} and \ref{universal}.

Now note that the union $L$ of the corresponding $L^{lk}$ 
is the list of tasks on which the SLIM NN's performance may have changed through the weight modifications.
All $T \notin L$ can be safely ignored---performance on those tasks remains unaffected.
For $T \in L$ 
we use {\bf Spread} to re-evaluate performance. 
If total performance on all $T \in L$  has not improved through the tentative weight changes, the latter are undone.
Otherwise we keep them, and
all affected $L^{pq}$ are updated as follows (using the $trace$s computed by {\bf Spread}): the new value $L^{pq}_i$ is obtained by appending to $L^{pq}_{i-1}$ those $T_j \notin L^{pq}_{i-1} (j=1,  \ldots, i)$ whose current (possibly revised) solutions now need  $w^{pq}$ at least once during the solution-computing process, and deleting those $T_j$ whose current solutions do not use  $w^{pq}$ any more.

That is, if the most recent task does not require changes of many weights, 
and if the changed weights do not affect many previous tasks, 
then validation of  learning progress 
through methods like those of Section \ref{universal} 
or similar  \cite{Schmidhuber:11powerplay}
may be much more efficient
than in traditional NN implementations.


\subsection{Additional LA Principles for SLIM NNs  on Future 3D  Hardware}
\label{3d}

Computers keep getting faster per cost.
To continue this trend within the  limitations of physics,
future hardware architectures will  feature
3-dimensional arrangements of numerous connected processors.
To minimize wire length and communication costs \cite{maass2002wire},
the processors should communicate through many low-cost
short-range and few high-cost long-range connections,
much like biological neurons.
Given some task,
to minimize energy consumption and cooling costs,
no more processors or neurons than necessary to solve the task should become active,
and those that communicate a lot with each other should typically  be physically close.

All of this can be encouraged through LAs
that punish excessive processing and communication costs of
3D SLIM NNs running on such hardware.

Consider the constant   $cost^{lk}$ of using $c^{lk}$ in such a 3D SLIM RNN
from one discrete time step to the next in {\bf Spread}-like procedures. 
$cost^{lk}$ may be viewed as the wire length of
$c^{lk}$  \cite{maass2002wire}.  
The expression $\sum_{l,k} cost^{lk} n(c^{lk})$ can enter the objective function, e.g., as an additive term
to be minimized by an LA like those mentioned in Section \ref{learning}. Note, however, that such costs are automatically 
taken into account by the universal program search methods of Section \ref{universal}.

Like biological brains, typical 3D SLIM RNNs will have many more
short wires than long ones.
An automatic by-product of LAs as in Section \ref{universal}
 should be the learning of subtask solutions by subsets of neurons
most of which are physically close to each other.
The resulting weight matrices may sometimes be reminiscent of self-organizing maps for pattern 
classification \cite{Kohonen:88} and motor control \cite{Graziano:book,ring:icdl2011}.
The underlying cause of such neighborhood-preserving weight matrices, 
however, will not be a traditional pre-wired 
neighborhood-enforcing learning rule \cite{Kohonen:88,ring:icdl2011},
but sheer efficiency {\em per se}.

\section{Experiments}
\label{experiments}
First experiments with SLIM NNs are currently conducted 
within the recent {\sc PowerPlay} framework \cite{Schmidhuber:11powerplay}.
 {\sc PowerPlay} is 
designed 
to learn a more and more general problem solver from scratch. 
The idea is to let a general computer (say, a SLIM RNN) solve more and more tasks 
from the infinite set of all computable tasks, 
without ever forgetting solutions to previously solved tasks.
At a given time, which task should be posed next?
Human teachers in general do not know which tasks are not yet solvable by the SLIM RNN through
generalization, yet easy to learn,
given what's already known. 
That's why {\sc PowerPlay} continually {\em invents the easiest-to-add new task by itself.}

To do this, {\sc PowerPlay} 
incrementally searches the space of possible pairs of (1) new tasks, 
and (2) SLIM RNN  modifications. 
The search continues until the first pair is discovered for which (a) the current SLIM RNN cannot solve the new task, and (b) the new SLIM RNN provably solves all previously learned tasks plus the new one. Here the new task may actually be to achieve a {\em wow-effect} by simplifying, compressing, or speeding up previous solutions. 

Given a SLIM RNN that can already solve a finite known set of previously learned tasks, 
a particular AOPS algorithm \cite{Schmidhuber:11powerplay} (compare Section \ref{universal})
can be used to find a new pair that provably has properties (a) and (b). Once such a pair is found, the cycle repeats itself. This results in a continually growing set of tasks solvable by an increasingly more powerful solver. The continually increasing repertoire of self-invented problem-solving procedures can be exploited at any time to solve externally posed tasks. 

How to represent tasks of the SLIM RNN? A unique task index is given as a constant RNN input in addition
to the changing inputs from the environment manipulated by the RNN outputs. 
Once the halting units gets activated and the computation ends,
the activations of  a special pre-defined set of internal neurons can be viewed as 
the result of the computation. Essentially
arbitrary computable tasks can be represented in this way by the SLIM RNN.

We can 
keep track of which tasks are dependent on each connection (Section \ref{tracking}).
If the most recent task to be learned 
does not require changes in many weights, and if the changed weights do not affect many previous tasks, 
then validation may be very efficient.
Now recall that 
{\sc PowerPlay} prefers to invent tasks whose validity check requires little computational effort. This implicit incentive (to generate modifications that do not impact many previous tasks), leads to a natural decomposition of the space of tasks and their solutions into more or less independent regions. Thus, divide and conquer strategies are
natural by-products of {\sc PowerPlay}-trained SLIM NNs \cite{Schmidhuber:11powerplay}.
Experimental results  will be reported in separate papers.

\section{Conclusion}
\label{conclusion}

Typical recurrent self-delimiting (SLIM) neural networks (NNs) are general 
computers for running arbitrary self-delimiting parallel-sequential programs encoded in their weights.
While certain types of SLIM NNs have been around for decades, e.g., \cite{Schmidhuber:89cs},
little attention has been given to certain fundamental benefits of their self-delimiting nature.
During program execution, lists or stacks can be used to trace only those neurons and connections used at least once.
This also allows for efficient resets of large NNs which  may use only a small fraction of their weights per task.
Efficient SLIM NN learning algorithms (LAs) track which weights are used for which tasks,
to greatly speed up performance evaluations in response to limited weight changes.
SLIM NNs are easily combined with techniques of asymptotically optimal program search.
To address overfitting, instead of depending on  pre-wired regularizers and hyper-parameters \cite{bishop:2006},
SLIM NNs  can in principle learn to select by themselves their own runtime and their own numbers of free parameters,
 becoming fast and  {\em slim} when necessary.
LAs may  penalize the task-specific  total length of connections used by 
SLIM NNs implemented on the 3-dimensional brain-like multi-processor hardware to expected in the future.
This should encourage SLIM NNs to solve many subtasks by subsets of neurons that are physically close.
Ongoing experiments with SLIM RNNs will be reported  separately.

\section{Acknowledgments}
\label{ack}
Thanks to Bas Steunebrink and Sohrob Kazerounian for useful comments.

\bibliography{bib}
\bibliographystyle{plain}
\end{document}